\documentclass[letterpaper, 10 pt, conference]{ieeeconf}  

\IEEEoverridecommandlockouts                              

\usepackage{graphics} 
\usepackage{epsfig} 
\usepackage{amsmath} 
\usepackage{amssymb}  
\usepackage{xcolor}
\usepackage{algorithm}
\usepackage{algpseudocode}
\usepackage{soul}

\newcommand{\subparagraph}{}
\usepackage{titlesec}
\usepackage{tikzpagenodes}

\newcommand\E{\mathbb{E}}

\title{\LARGE \bf
Variational End-to-End Navigation and Localization 
}

\author{Alexander Amini$^{1}$, Guy Rosman$^{2}$, Sertac Karaman$^{3}$ and Daniela Rus$^{1}$
\thanks{Support for this work was given by the National Science Foundation (NSF) and Toyota Research Institute (TRI). We gratefully acknowledge the support of NVIDIA Corporation with the donation of the V100 GPU and Drive PX2 used for this research.}
\thanks{$^{1}$ Computer Science and Artificial Intelligence Lab, Massachusetts Institute of Technology \{amini,rus\}@mit.edu}
\thanks{$^{2}$ Toyota Research Institute \{guy.rosman\}@tri.global}
\thanks{$^{3}$ Laboratory for Information and Decision Systems, Massachusetts Institute of Technology \{sertac\}@mit.edu}}

\begin{document}

\maketitle


\renewcommand{\baselinestretch}{0.97}

\thispagestyle{empty}
\pagestyle{empty}

%

\begin{abstract}

Deep learning has revolutionized the ability to learn ``end-to-end'' autonomous vehicle control directly from raw sensory data. 
%
While there have been recent extensions to handle forms of navigation instruction, these works are unable to capture the full distribution of possible actions that could be taken and to reason about localization of the robot within the environment.
In this paper, we extend end-to-end driving networks with the ability to perform point-to-point navigation as well as probabilistic localization using only noisy GPS data. 
We define a novel variational network capable of learning from raw camera data of the environment as well as higher level roadmaps to predict (1) a full probability distribution over the possible control commands; and (2) a deterministic control command capable of navigating on the route specified within the map. 
Additionally, we formulate how our model can be used to localize the robot according to correspondences between the map and the observed visual road topology, inspired by the rough localization that human drivers can perform. 
We test our algorithms on real-world driving data that the vehicle has never driven through before, and integrate our point-to-point navigation algorithms onboard a full-scale autonomous vehicle for real-time performance. 
Our localization algorithm is also evaluated over a new set of roads and intersections to demonstrates rough pose localization even in situations without any GPS prior.
%



\end{abstract}

\section{Introduction}

Human drivers have an innate ability to reason about the high-level structure of their driving environment even under severely limited observation. They use this ability to relate high-level driving instructions to concrete control commands, as well as to better localize themselves even without concrete localization information. Inspired by these abilities, we develop a learning engine that enables a robot vehicle to learn how to use maps within an end-to-end autonomous driving system.

Coarse grained maps afford us a higher-level of understanding of the environment, both because of their expanded scope, but also due to their distilled nature. This allows for reasoning about the low-level control within a hierarchical framework with long-term goals \cite{sutton1999between}, as well as localizing, preventing drift, and performing loop-closure when possible. We note that unlike much of the work in place recognition and loop closure, we are looking at a higher level of matching, where the vehicle is matching intersection and road patterns to the coarse scale geometry found in the map, allowing handling of different appearances and small variations in the scene structure, or even unknown fine-scale geometry, as long as the overall road network structure matches the expected structures.

While end-to-end driving ~\cite{bojarski2016end} holds promise due to its easily scalable and adaptable nature, it has a limited capability to handle long-term plans, relating to the nature of imitation learning  \cite{codevilla2017end, shalev2016safe}. Some recent methods incorporate maps as inputs \cite{wei2017intention,hecker2018end} to capture longer term action structure, yet they ignore the uncertainty maps inherently allow us to address -- uncertainty about the location, and uncertainty about the longer-term plan. 

\begin{figure}[t!]
    \centering
    \includegraphics[width=1\linewidth]{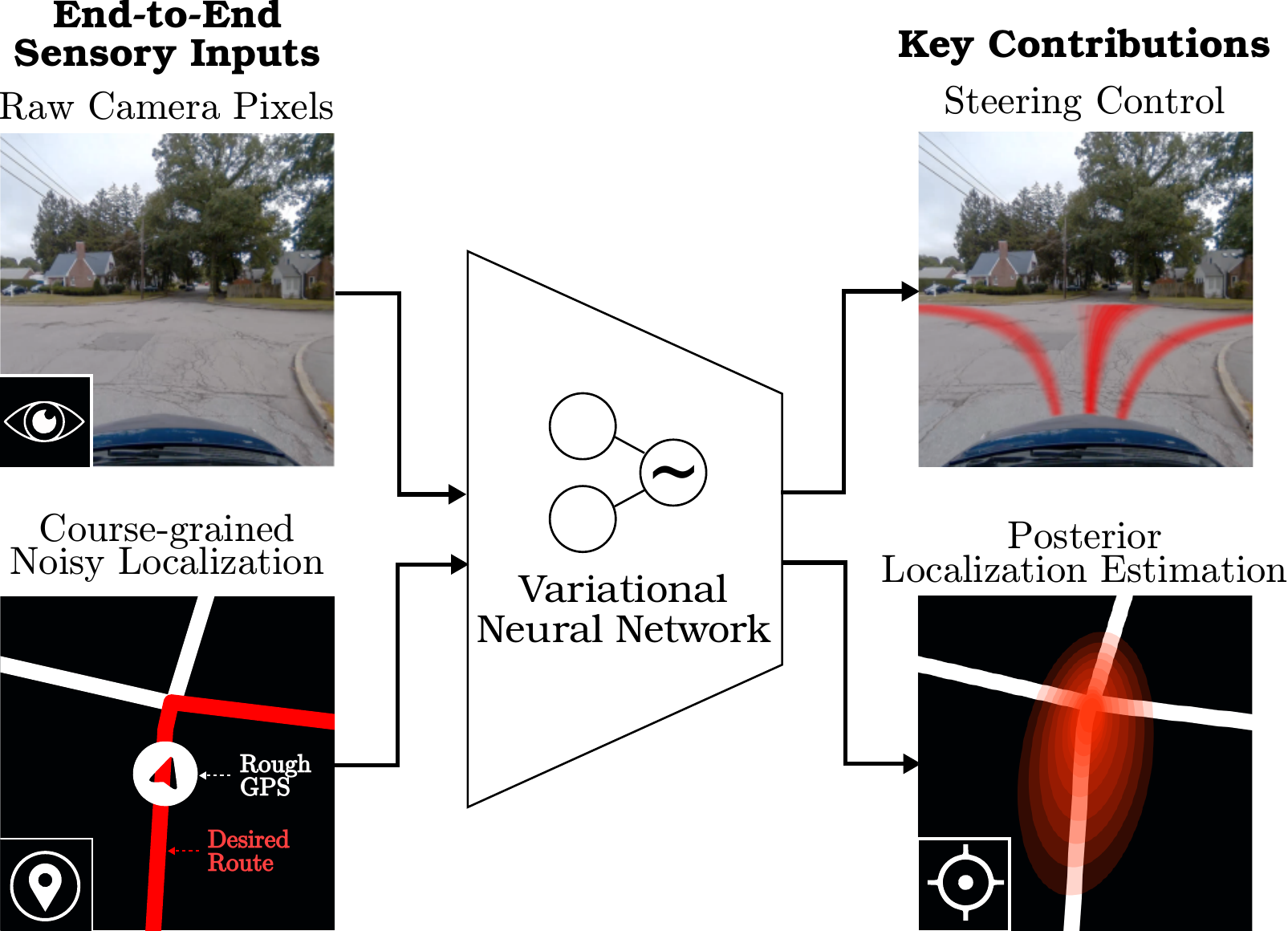}
    \caption{\textbf{Variational end-to-end model.} Our model learns from raw sensory data as well as coarse grained topology maps to navigate and localize within complex environments.}
    \label{fig:intro}
\end{figure}

\begin{figure*}[t!]
    \centering
    \includegraphics[width=1\textwidth]{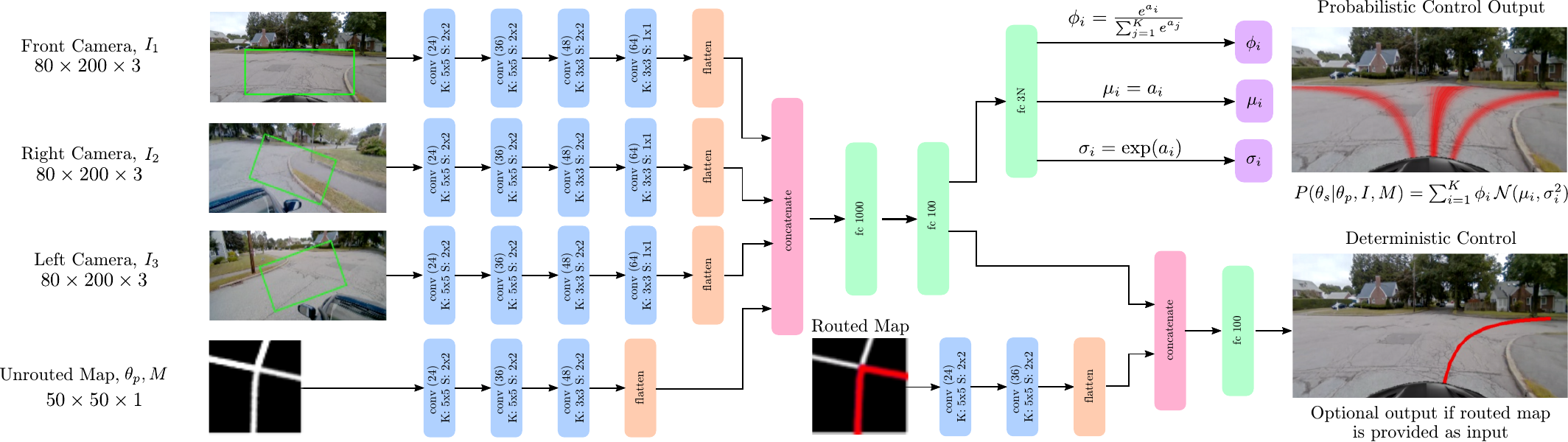}
    \caption{\textbf{Model architecture overview.} Raw camera images and noisy roadmaps are fed to parallel convolutional pipelines, then merged into fully-connected layers to learn a full parametric Gaussian Mixture Model (GMM) over control. If a routed map is also available, it is merged at the penultimate layer to learn a deterministic control signal for navigation along a provided route. Green rectangles denote the image region provided as input to the network.  }
    \label{fig:model}
\end{figure*}

In this paper, we address these limitations by developing a novel model for integrating navigational information with raw sensory data into a single end-to-end variational network, and do so in a way that preserves reasoning about uncertainty. This allows the system to not only learn to navigate complex environments entirely from human perception and navigation data, but also understand when localization or mapping is incorrect, and thus correct for the pose (cf. Fig.~\ref{fig:intro}).

Our model processes coarse grained, unrouted roadmaps, along with forward facing camera images to produce a probabilistic estimate of the different possible low-level steering commands which the robot can execute at that instant. In addition, if a routed version of the same map is also provided as input, our model has the ability to output a deterministic steering control signal to navigate along that given route. 

The key contributions of this paper are as follows:
\begin{itemize}
    \item Design of a novel variational end-to-end control network which integrates raw sensory data with routed and unrouted maps, enabling navigation and localization in complex driving environments; and
    \item Formulation of a localization algorithm using our trained end-to-end network to reason about a given road topology and infer the robot's pose by drawing correspondences between the map and the visual road appearance; and
    \item Evaluation of our algorithms on a challenging real-world dataset, demonstrating navigation with steering control as well as improved pose localization even in situations with severely limited GPS information.
\end{itemize}

The remainder of the paper is structured as follows: we summarize the related work in Sec.~\ref{sec:related}, formulate the model and algorithm for posterior pose estimation in Sec.~\ref{sec:model},  describe our experimental setup, dataset, and results in Sec.~\ref{sec:results}, and provide concluding remarks in Sec.~\ref{sec:conclusion}.

\section{Related Work}
\label{sec:related}

Our work ties in to several related efforts in both control and localization.
As opposed to traditional methods for autonomous driving which typically rely on distinct algorithms for localization and mapping~\cite{leonard1991simultaneous,montemerlo2002fastslam,davison2007monoslam}, planning~\cite{kavraki1994probabilistic,lavalle2000rapidly,karaman11}, and control~\cite{schwarting2017safe,falcone2007predictive}, end-to-end algorithms attempt to collapse the problem (directly from raw sensory data to output control commands) into a single learned model. The ALVINN system~\cite{pomerleau1989alvinn} originally proposed the use of multilayer perceptron to learn the direction a vehicle should steer in 1989. Recent advancements in convolutional neural networks (CNNs) have revolutionized the ability to learn, directly from raw imagery, either a deterministic~\cite{bojarski2016end}, or probabilistic ~\cite{amini2018learning,amini_nips_2017} driving command (i.e. steering wheel angle or road curvature). Followup works have incorporated conditioning on additional cues \cite{codevilla2017end,huang2019uncertainty}, including mapped information \cite{wei2017intention,hecker2018end}.  However, these works do not relate the the uncertainty of multiple steering possibilities to the map, nor do they present the ability to reason about discrepancy between their input modalities.

A recent line of work has tied end-to-end driving networks to variational inference \cite{amini2018variational}, allowing us to handle cases where multiple actions are possible, as well as reason about robustness, atypical data, and dataset normalization. Our work extends this line and allows us to use the same outlook to reason about maps as an additional conditioning factor.

Our work also relates to several research efforts in reinforcement learning in subfields such as bridging different levels of planning hierarchies \cite{sutton1999between,fox2017multi}, and relating to maps as agents plan and act \cite{tamar2016value,oh2017value}. This work relates to a vast literature in localization and mapping \cite{leonard1991simultaneous,montemerlo2002fastslam}, such as visual SLAM \cite{davison2007monoslam}, \cite{engel2014lsd} and place recognition \cite{paul2010fab,sattler2011fast}. However, our notion of visual matching is much more high-level, more akin to semantic visual localization and SLAM \cite{schonberger2018semantic,bowman2017probabilistic}, where the semantic-level features are driving affordances.

\section{Model}
\label{sec:model}

\begin{figure*}
    \centering
    \includegraphics[width=1\linewidth]{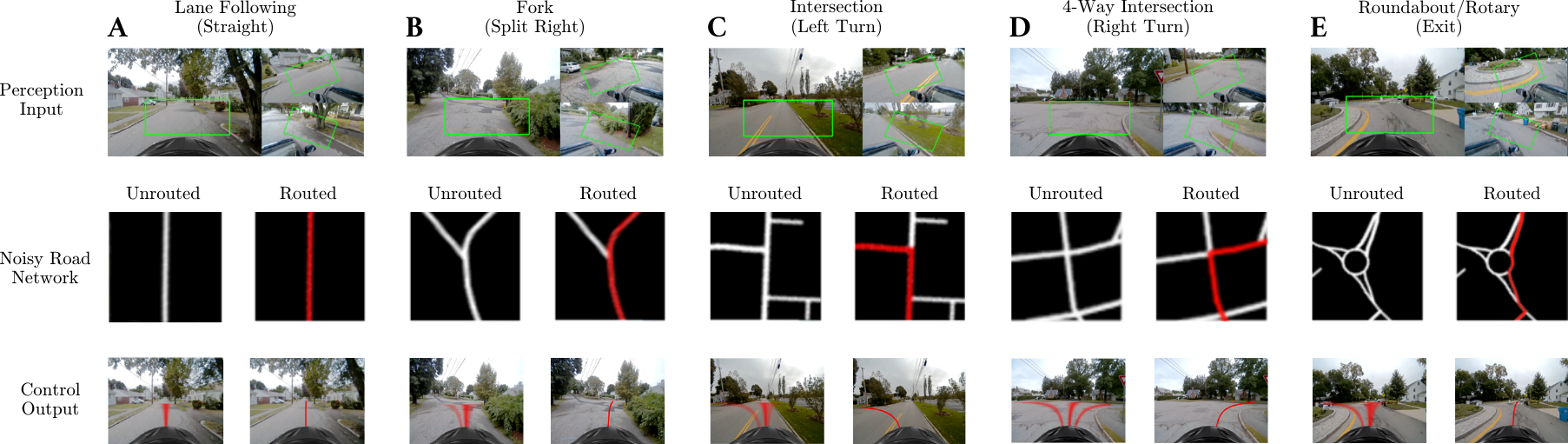}
    \caption{\textbf{Control output under new, rich road environments.} Demonstration of our system which takes as input (A) image perception (green box denotes patch fed to the model); and (B) coarse unrouted roadmap (and routed, if available). The output (C) is a full continuous probability distribution for unrouted maps, and a deterministic command for navigating on a routed map. We demonstrate the control output on five scenarios, of roughly increasing complexity (left to right), ranging from straight road driving, to intersections, and even a roundabout. }
    \label{fig:driving_example}
\end{figure*}


In this section, we describe the model used in our approach. We use a variational neural network, which takes raw camera images, $I$, and an image of a noisy, unrouted roadmap, $M_U$, as input. At the output we attempt to learn a full, parametric probability distribution over road curvature or steering ($\theta_s$) to navigate that instant. We use a Gaussian Mixture Model (GMM) with $K>0$ modes to describe the possible steering control command, and penalize the $L_{1/2}$ norm of the weights to discourage extra components. Empirically, we chose $K=3$ since it captured the majority of driving situations encountered. Additionally, the model will optionally output a deterministic control command if a routed version of the map is also provided as input. The overall network can be written separately as two functions, representing the stochastic (unrouted) and deterministic (routed) parts respectively:  
\begin{align}
    \{(\phi_i,\mu_i,\sigma_i^2)\}_{i=1}^K = f_S(I,M_U,\theta_p),\\
    \nonumber \hat \theta_s = f_D(I,M_R,\theta_p),
\end{align}
where $\theta_p = [p_x, p_y, p_\alpha]$ is the current pose in the map (position and heading), and $f_S(I,M_U,\theta_p), \, f_D(I,M_R,\theta_p)$ are network outputs computed by cropping a square region of the relevant map according to $\theta_p$, and feeding it, along with the forward facing images, $I$, to the network. $M_U$ denotes the \emph{unrouted} map, with only the traversible areas marked, while $M_R$ denotes the \emph{routed} map, containing the desired route highlighted. The deterministic control command is denoted as $\hat\theta_s$. In this paper, we refer to steering command interchangeably as the road curvature: the actual steering angle requires reasoning about road slip and control plant parameters that change between vehicles, making it less suitable for our purpose. Finally, the parameters (i.e. weight, mean, and variance) of the GMM's $i$-th component are denoted by $(\phi_i,\mu_i,\sigma_i^2)$, which represents the steering control in the absence of a given route.

The overall network structure is given in Fig.~\ref{fig:model}. Each camera image is processed by a separate convolutional pipeline similar to the one used in \cite{bojarski2016end}. 
Similarly, the cropped, non-routed, map patch is fed to a set of convolutional layers, before concatenation to the image processing outputs. However, here we use fewer layers for two main reasons: a) The map images contain significantly fewer features and thus don't require a complex feature extraction pipeline and b) We wish to avoid translational invariance effects often associated with convolutional layers and subsampling, as we are interested in the pose on the map. The output of the convolutional layers is flattened and fed to a set of fully connected layers to produce the parameters of a probability distribution of steering commands (forming $f_S$). As a second task, we the previous layer output along with a convolutional module processing the routed map, $M_R$, to output a single deterministic steering command, forming $f_D$. This network structure allows us to handle both routed and non-routed maps, and later affords localization and driver intent, as well as driving according to high level navigation (i.e. turn-by-turn instruction).

We learn the weights of our model using backpropogation with the loss defined as:
\begin{align}
    \E\left\{\begin{matrix}
    \mathcal{L}\Big(f_S(I,M,\theta_p),\theta_s\Big)+\|\phi\|_{p}+\\ 
    \sum_i \psi_{S}(\sigma_i) + \Big(f_D(I,M,\theta_p)-
    \theta_s\Big)^2\end{matrix}\right\}
\end{align}

where $\psi_{S}$ is a per-component penalty on the standard deviation $\sigma_i$. We chose a quadratic term in log-$\sigma$ as the regularization,
\begin{align}
    \psi_{S}(\sigma) = \|\log \sigma - c\|^2.
\end{align}
$\mathcal{L}\big(f_S(I,M,\theta_p,),\theta_s\big)$ is the negative log-likelihood of the steering command according to a GMM with parameters $\{(\phi_i,\mu_i,\sigma_i)\}_{i=0}^N$ and
\begin{align}
    P(\theta_s|\theta_p,I,M) = \sum \phi_i \mathcal{N}(\mu_i,\sigma_i^2).
\end{align}

\subsection{Localization via End-to-End Networks}

The conditional structure of the model affords updating a posterior belief about the vehicle's pose, based on the relation between the map and the road topology seen from the vehicle. For example, if the network is provided visual input, $I$, which appears to be taken at a 4 way intersection we aim to compute $P(\theta_p | I,M)$ over different poses on the map to reason about where this input could have been taken. Note that our network only computes $P(\theta_s |\theta_p, I, M)$, but we are able to estimate our pose given the visual input through double marginalization over $\theta_s$ and $\theta_p$. Given a prior belief about the pose, $P(\theta_p)$, we can write the posterior belief after seeing an image, $I$, as: 
\begin{align}
\label{eq:posterior_probability}
    \nonumber P(\theta_p|I,M) &= \E_{\theta_s} P(\theta_p|\theta_s,I,M) \\
    &= \E_{\theta_s} \left[ \frac{P(\theta_p,\theta_s|I,M)}{P(\theta_s|I,M)} \right]\\
    &= \E_{\theta_s} \left[ \frac{P(\theta_p,\theta_s|I,M)}{\E_{\theta_{p'} }P(\theta_s|\theta_{p'},I,M)} \right]& \nonumber\\
    &= \E_{\theta_s} \left[ \frac{P(\theta_s|\theta_p,I,M)}{\E_{\theta_{p'}} P(\theta_s|\theta_{p'},I,M)} \, P(\theta_p) \right],  \nonumber
\end{align}
where the equalities are due to full probability theorem and Bayes theorem.
The posterior belief can be therefore computed via marginalization over $\theta_p,\theta_s$. While marginalization over two random variables is traditionally inconvenient, in two cases of interest, marginalizing over $\theta_p$ becomes easily tractable: a) when the pose is highly localized due to previous observations, as in the case of online localization and b) where the pose is sampled over a discrete road network, as is done in mapmatching algorithms. The algorithm to update the posterior belief is shown in Algorithm~\ref{alg:posterior_update}. Intuitively, the algorithm computes, over all steering angle samples, the probability that a specific pose and images/map explain that steering angle, with the additional loop required to estimate the partition function and normalize the distribution. We note the same algorithm can be used with small modifications within the map-matching framework~\cite{bernstein1998introduction, newson2009hidden}.

\begin{algorithm}[ht]
\caption{Posterior Pose Estimate from Driving Direction\label{alg:posterior_update}}
\textbf{Input:} $I,\, M,\, p(\theta_p)$\\
\textbf{Output:} $P(\theta_p | I,M)$
\begin{algorithmic}
\For {$i=1...N_{s}$:} 
\State Sample $\theta_s$
\State Compute $P(\theta_s|\theta_p,I,M)$
\For {$j=1...N_{p}$:} 
\State Compute $P(\theta_s|\theta_{p'},I,M)$
\State Aggregate $\E_{\theta_{p'}} P(\theta_s|\theta_{p'},I,M)$
\EndFor

Aggregate $\E_{\theta_s} \left[ \frac{P(\theta_s|\theta_p,I,M)}{\E_{\theta_{p'} }P(\theta_s|\theta_{p'},I,M)} \, P(\theta_p) \right]$
\EndFor
\State Output $P(\theta_p | I,M)$ according to Equation~\ref{eq:posterior_probability}.
\end{algorithmic}
\end{algorithm}

\section{results}
\label{sec:results}

In this section, we demonstrate results obtained using our method on a real-world train and test dataset of rich driving enviornments. We start by describing our system setup and dataset and then proceed to demonstrate driving using an image of the roadmap and steering angle estimation with both routed and unrouted maps. Finally, we demonstrate how our approach allows us to reduce pose uncertainty based on the agreement between the map and the camera feeds.

\subsection{System Setup}
\label{subsec:system_setup}
We evaluate our system on a 2015 Toyota Prius V outfitted with autonomous drive-by-wire capabilities~\cite{naserIV2017}. Additionally, we made several advancements to the sensor and compute platform specifically for this work. Three Leopard Imaging LI-AR0231-GMSL cameras~\cite{LeopardImaging0231Datasheet}, capable of capturing 1080p RGB images at approximately 30Hz, are used as the vision data source for this study. We mount the three cameras on the front of the vehicle at various yaw angles: one forward facing and the remaining two rotated on the left/right of the vehicle to capture a larger FOV. Coarse grained global localization is captured using the OXTS RT3000 GPS~\cite{OXTSUserManual} along with an Xsense MTi 100-series IMU~\cite{XsenseWhitePaper}. We use the yaw rate $\gamma$ [\texttt{rad/sec}], and the speed of the vehicle, $v$  [\texttt{m/sec}], to compute the curvature (or inverse steering radius) of the path which the human executed as $\theta_s = \frac{\gamma}{v}$. Finally, all of the sensor processing was done onboard an NVIDIA Drive PX2~\cite{NvidiaPX2Docs}.

In order to build the road network, we gather edge information from Open Street Maps (OSM) throughout the traversed region. We are given a directed topological graph, $G(V,E)$, where $v_i \in V$ represents an intersection on the road network and $e_i\in E$ represents a single directed road between two intersections. The weight of every edge, $w(e_i)$, is defined according to its great circle length, but with slight modifications could also capture more complexity by incorporating the type of road or even real-time traffic delays. The problem of offline map-matching 
involves going from a noisy set of ordered poses, $\{\theta_p^{(t)}\}_{t=1}^N$, to corresponding set of traversed road segments $\{e_i^{(t)}\}_{t=1}^N$ (i.e. the route taken). We implemented our map matching algorithm as an offline pre-processing step as in~\cite{newson2009hidden}. 

One concern during map rendering process is handling \emph{ambiguous} parts of the map. Ambiguous parts are defined as parts where the the driven route cannot be described as a single simple curve. In order to handle large scale driving sequences efficiently, and avoid map patches that are ambiguous, we break the route into non-ambiguous subroutes, and generate the routed map for each of the subroutes, forming a set of charts for the map. For example, in situations where the vehicle travels over the same intersection multiple times but in different directions, we should split the route into distinct segments such that there are no self-crossings in the rendered route. Finally, we render the unrouted map by drawing all edges on a black canvas. The routed map is rendered by adding a red channel to the canvas and by drawing the traversed edges, $\{e_i^{(t)}\}_{t=1}^N$. 

Using the rendered maps and raw perceptual image data from the three cameras, we trained our model (implemented in TensorFlow~\cite{abadi2016tensorflow}) using 25 km of driving data taken in a suburban area with various different types of turns, intersections, roundabouts, as well as other dynamic obstacles (vehicles and pedestrians). We test and evaluate our results on an entirely different set of road segments and intersections which the network was never trained on. 

\subsection{Driving with Navigational Inputs}
We demonstrate the ability of the network to compute both the continuous probability distribution over steering control for a given unrouted map as well as the deterministic control to navigate on a routed map. In order to drive with a navigational input, we feed both the unrouted and the routed maps into the network and compute $f_D(I,M,\theta_p)$.
In Fig.~\ref{fig:driving_example} we show the inputs and parametric distributions of steering angles of our system. The roads in both the routed and unrouted maps are shown in white. In the routed map, the desired path is painted in red. In order to generate the trajectories shown in the figure, we project the instantaneous control curvature as an arc, into the image frame. In this sense, we visualize the projected path of the vehicle if it was to execute the given steering command. Green rectangles in camera imagery denote the region of interest (ROI) which is actually actually fed to the network. We crop according to these ROIs so the network is not provided extra non-essential data which should not contribute to its control decision (e.g., the pixels above the horizon line do not impact steering). 

We visualize the model output under various different driving scenarios ranging (left to right) from simple lane following to richer situations such as turns and intersections (cf. Fig.~\ref{fig:driving_example}). In the case of lane following (left), the network is able to identify that the GMM only requires a single Gaussian mode for control, whereas multiple Gaussian modes start to appear for forks, turns, and intersections. In the case where a routed path is also provided, the network is able to disambiguate from the multiple modes and select a correct control command to navigate towards the route. We also demonstrate generalization on a richer type of intersection such as the roundabout (right) which was never included as part of the training data. Furthermore, we integrate our proposed end-to-end navigation stack onboard our full-scale autonomous vehicle~\cite{naserIV2017} for real-time performance (15Hz) through an unseen test track spanning approximately 1 km (containing a total of 9 intersections). 

\begin{figure}[t!]
    \centering
    \includegraphics[width=1\linewidth]{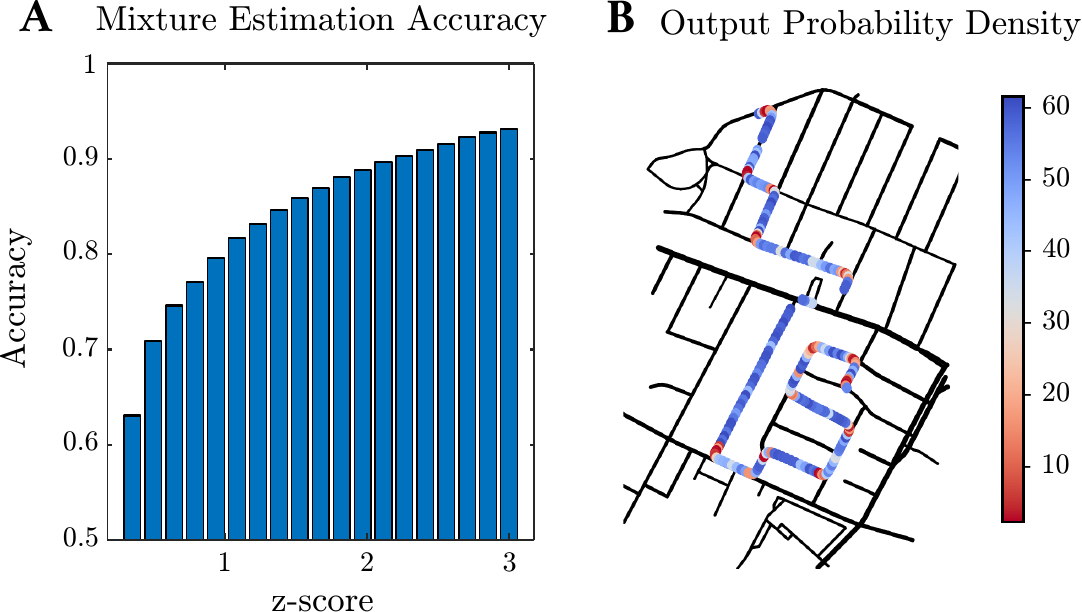}
    \caption{\textbf{Model evaluation.} (A) The fraction of the instances when the true (human) steering command was within a certain z-score of our probabilistic network. (B) The probability density of the true steering command as a function of spatial location in the test roadmap. As expected the density decreases before intersections as the overall measure is divided between multiple paths. Points with gross GPS failures were omitted from visualization. }
    \label{fig:mixture_acc}
\end{figure}

To quantitatively evaluate our network we compute the mixture estimation accuracy over our entire test set (cf. Fig~\ref{fig:mixture_acc}). Specifically, for a range of z-scores over the steering control distribution we compute the number of samples within the test set where the true (human) control output was within the predicted range. To provide a more qualitative understanding of the spatial accuracy of our variational model we also visualized a heatmap of GPS points over the test set (cf. Fig.~\ref{fig:mixture_acc}B), where the color represents the probability density of the predicted distribution evaluated at the true control value. We observed that the density decreases before intersections as the modes of the GMM naturally spread out to cover the greater number of possible paths. 

\subsection{Reducing Localization Uncertainty}

We demonstrate how our model can be used to localize the vehicle based on the observed driving directions using Algorithm~\ref{alg:posterior_update}. We investigate in our experiments the reduction of pose uncertainty, and visualize areas which offer better types of pose localization. 

\begin{figure}[t!]
    \centering
    \includegraphics[width=1\linewidth]{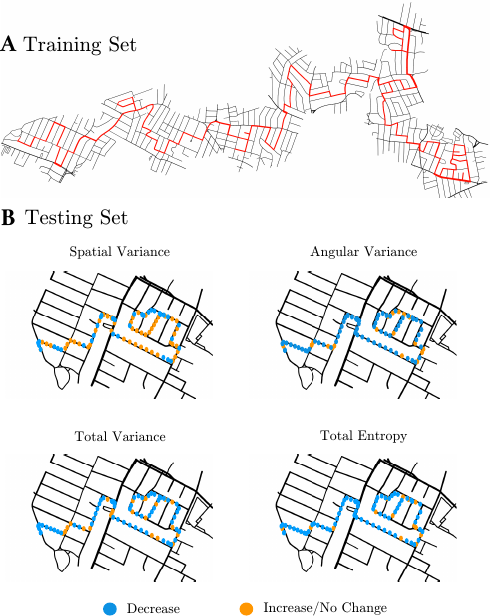}
    \caption{\textbf{Evaluation of posterior uncertainty improvement.} (A) A roadmap of the data used for training with the route driven in red (total distance of 25km). (B) A heatmap of how our approach increases/decreases four different types of variance throughout test set route. Points represent individual GPS readings, while the color (orange/blue) denotes the absolute impact (increase/decrease) our algorithm had on its respective variance. Decreasing variance (i.e. increasing confidence) is the desired impact of our algorithm.  
    }
    \label{fig:map_heatmaps}
\end{figure}

For this experiment, we began with the pose obtained from the GPS and assumed an initial error in this pose with some uncertainty (Gaussian over the spatial position, heading, or both). We compute the posterior probability of the pose as given by Alg.~\ref{alg:posterior_update}, and look at the individual uncertainty measures or total entropy of the prior and posterior distributions. If the uncertainty in the posterior distribution is lower than that of the prior distribution we can conclude that our learned model was able to increase its localization confidence after  seeing the visual inputs provided (i.e. the camera images). In Fig.~\ref{fig:map_heatmaps} we show the training (A) and testing (B) datasets. Note that the roads and intersections in both of these datasets were entirely disjoint; the model was never trained on roads/intersections from the test set.  

For the test set, we overlaid the individual GPS points on the map and colored each point according to whether our algorithm increased (blue) or decreased (orange) posterior uncertainty. When looking at uncertainty reduction, it is important to note which degrees of freedom (i.e. spatial vs angular heading) localize better at different areas in the road network. For this reason, we visualize the uncertainty reduction heatmaps four times individually across (1) spatial variance, (2) angular variance, (3) overall pose variance, and (4) overall entropy reduction (cf. Fig.~\ref{fig:map_heatmaps}). 

While header angle is corrected easily at both straight driving and more complex areas (turns and intersections), spatial degrees of freedom are corrected best at rich map areas, and poorly at linear road segments. This is expected and is similar to the aperture problem in computer vision \cite{marr1981directional} -- the information in a linear road geometry is not enough to establish 3DOF localization.

If we focus on areas preceding intersections (approx 20 meters before), we typically see that the spatial uncertainty (prior uncertainty of $2m$) is reduced right before the intersection, which makes sense since after we pass through our forward facing visual inputs are not able to capture the intersection behind the vehicle. Looking in the vicinity of intersections, we achieved average reduction of $0.31$ nats. For the angular uncertainty, with initial uncertainty of $\sigma=0.8$ radians ($45$ degs), we achieved a reduction in the standard deviation of 0.2 radians (11 degs). 

\begin{figure}[t!]
    \centering
    \includegraphics[width=1\linewidth]{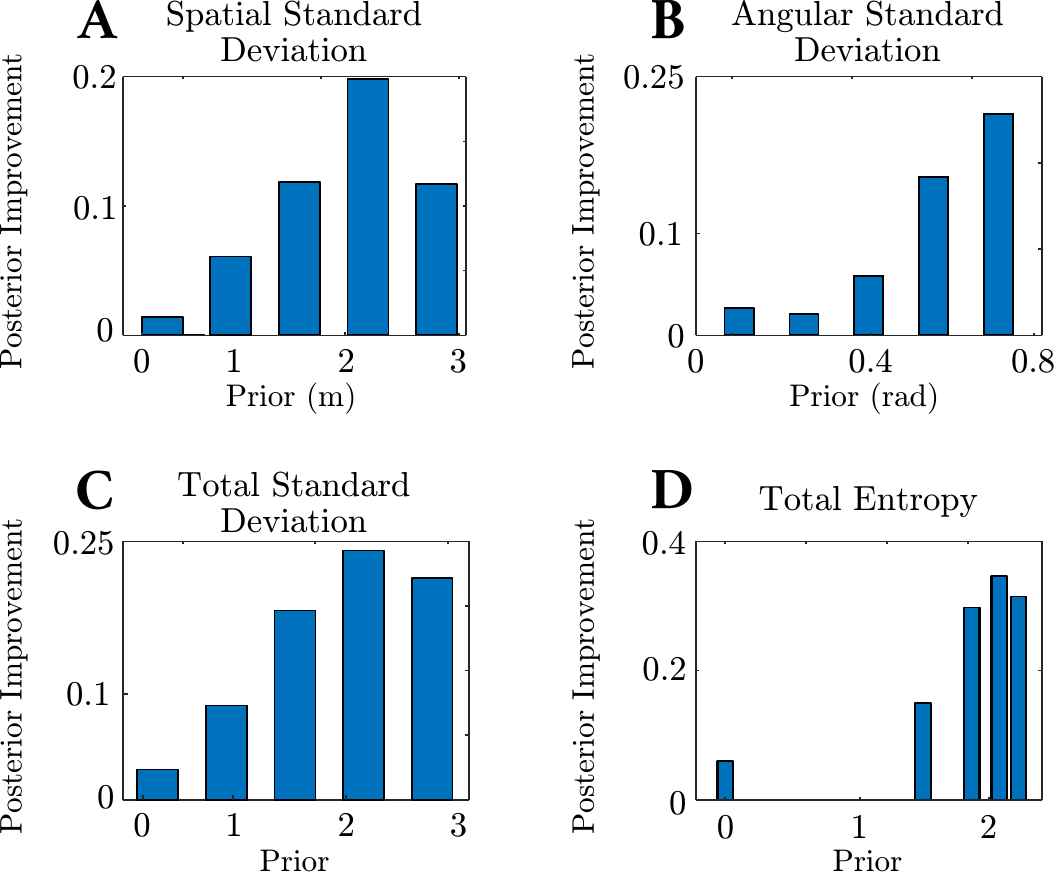}
    \caption{\textbf{Pose uncertainty reduction at intersections.} The reduction of uncertainty in our estimated posterior across varying levels of added prior uncertainty. We demonstrate improvement in A) spatial $\sigma^2(p_x)+\sigma^2(p_y)$, B) angular: $\sigma^2(p_\alpha)$, C) sum of variance over $p_x,p_y,p_\alpha$, and D) entropy in $p_x,p_y,p_\alpha$, Gaussian approximation. Note that we observe a ``positive'' improvement over all levels of prior uncertainty (averaged over all samples in regions preceding intersections). } 
    \label{fig:posterior_plots}
\end{figure}

We quantify the degree of posterior uncertainty reduction around intersections in Fig.~\ref{fig:posterior_plots}. Specifically, for each of the degrees of uncertainty (spatial, angular, etc) in Fig.~\ref{fig:map_heatmaps} we present the corresponding numerical uncertainty reduction as a function of the prior uncertainty in Fig.~\ref{fig:posterior_plots}. Note that we obtain reduction of both heading and spatial uncertainty for a variety of prior uncertainty values. Additionally, the averaged improvement over intersection regions is always positive for all prior uncertainty values indicating that localization does not worsen (on average) after using our algorithm.

\subsection{Coarse Grained Localization}

\begin{figure}[t!]
    \centering
    \includegraphics[width=1\linewidth]{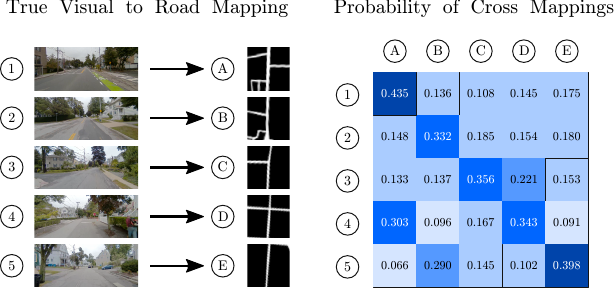}
    \caption{\textbf{Coarse Localization from Perception.} Five example locations from the test set (image, roadmap pairs). Given images from location $i$, we compute the network's probability conditioned on map patch from location $j$ in the confusion matrix. Thus, we demonstrate how our system can establish correspondences between its camera and map input and even determine when its map pose has a gross error. }
    \label{fig:inter_confusion}
\end{figure}

We next evaluate our model's ability to distinguish between significantly different locations without any prior on pose. For example, imagine that you are in a location without GPS but still want to perform rough localization given your visual surroundings (similar to the kidnapped robot problem). We seek to establish correspondences between the map and the visual road area for coarse grained place recognition. 

In Fig.~\ref{fig:inter_confusion} we demonstrate how we can identify and disambiguate a small set of locations, based on the the map and the camera images' interpreted steering direction. Our results show that we can easily distinguish between places of different road topology or road geometry, in a way that should be invariant to the appearance of the region or environmental conditions. Additionally, the cases where the network struggles to disambiguate various poses is understandable. For example, when trying to determine which map the image from environment 4 was taken, the network selects maps A and D where both have upcoming left and right turns. Likewise, when trying to determine the location of environment 5, maps B and E achieve the highest probabilities. Even though the road does not contain any immediate turns, it contains a large driveway on the lefthand side which resembles a possible left turn (thus, justifying the choice of map B). However, the network is able to correctly localize each of these five cases to the correct map location (i.e. noted by the strong diagonal of the confusion matrix).

\section{Conclusion}
\label{sec:conclusion}
In this paper, we developed a novel variational model for incorporating coarse grained roadmaps with raw perceptual data to directly learn steering control of autonomous vehicle. We demonstrate deterministic prediction of control according to a routed map, estimation of the likelihood of different possible control commands, as well as localization correction and place recognition based on the map. We formulate a concrete pose estimation algorithm using our learned network to reason about the localization of the robot within the environment and demonstrate reduced uncertainty (greater confidence) in our resulting pose. 

In the future, we intend to also integrate our localization algorithm in the \emph{online} setting of discrete road map-matching onboard our full-scale autonomous vehicle and additionally provide a more robust evaluation of the localization compared to that of a human driver. 

\bibliographystyle{IEEEtran} {
    \bibliography{ref}  
}

\end{document}